%% file: main.tex
\documentclass[11pt]{article}

\usepackage[final]{acl}

\usepackage{times}
\usepackage{latexsym}

\usepackage[T1]{fontenc}

\usepackage[utf8]{inputenc}

\usepackage{microtype}

\usepackage{inconsolata}

\usepackage{graphicx}

\usepackage{times}
\usepackage{latexsym}
\usepackage{amsmath}
\usepackage{amssymb} 
\usepackage{booktabs} 
\usepackage[most]{tcolorbox}
\newtcolorbox{qualex}[1]{
  colback=gray!4, colframe=gray!35, boxrule=0.4pt,
  left=4pt, right=4pt, top=3pt, bottom=3pt,
  title=\textit{#1}, fonttitle=\footnotesize,
  coltitle=black, colbacktitle=gray!12,
  breakable
}
\usepackage{multirow}

\title{A Coin Flip Per Token: Bernoulli Sparse Steering of Large Language Models}

\author{
Nima Eshraghi$^1$, Lovedeep Gondara$^1$, Yuqing Huang$^1$, Sagarika Suresh$^1$, \\
\textbf{Leizer Teran$^1$, Jithin Pradeep$^1$, Xiaotong (Tone) Xu$^2$, Fanny Chevalier$^2$} \\
$^1$The Vanguard Group, Inc. 
 $^2$University of Toronto \\
{\small\texttt{\{nima\_eshraghi, lovedeep\_gondara, yuqing\_huang, sagarika\_thimmanayakanapalya, }} \\
{\small\texttt{leizer\_teran, jithin\_pradeep\}@vanguard.com}} \\
{\small\texttt{\{tonexu, fanny\}@cs.toronto.edu}}
}

\begin{document}
\maketitle
\begin{abstract}
Activation steering via sparse autoencoders (SAEs) enables behavioral control of large language models without task-specific fine-tuning, but standard methods apply the steering signal at every generated token, incurring constant per-token perturbation that risks degrading fluency. We ask: \emph{is dense intervention necessary?} We introduce Stochastic Token Steering (STS), which gates each token independently with probability $p$, and Stochastic Block Steering (SBS), which gates a leading window once per sequence; neither requires a reward model or learned gating policy. Across two model families and two behavioral tasks, steering only a fraction of tokens recovers most of the behavioral effect of dense steering while incurring less fluency degradation. Across our experimental settings, we further find that the best-performing steering magnitude consistently rises as the intervention ratio falls.

\end{abstract}

\section{Introduction}
Behavioral control over large language models (LLMs), reducing toxicity, enforcing a
persona, adjusting refusal, is a significant deployment
concern~\citep{ouyang2022training,bai2022constitutional}. The default mechanism is the
system prompt, but prompt-based steering is fragile: output distributions shift under
minor lexical perturbation, adversarial suffixes override alignment priors, and
steerability through prompting is asymmetric and
bounded~\citep{mizrahi2024state,wei2023jailbroken,zou2023universal,wolf2023fundamental}.

Activation steering modifies model behavior by injecting a steering direction into
an intermediate residual stream during generation
\citep{turner2025steering,li2023inference,rimsky2024steering,
zou2023representation}. Sparse Autoencoders (SAEs) provide an interpretable basis for
such interventions by decomposing activations into sparse features whose decoder
directions can be amplified to promote or suppress a target behavior
\citep{cunningham2023sparse,bricken2023monosemanticity,
labonne2024eiffel,templeton2024scaling}.

Standard activation-steering implementations typically apply the selected direction
at every generated token. This design leaves open whether persistent intervention is
necessary, particularly because stronger or more frequent intervention can degrade
generation quality \citep{turner2025steering,labonne2024eiffel}. We therefore ask:
\emph{how much of the behavioral effect of dense steering remains when only a
fraction of token positions are steered?} We study this question by replacing
deterministic every-token intervention with stochastic gating and measuring the
behavioral and fluency effects across intervention ratios.

We introduce two stochastic interventions over SAE feature steering. \emph{Stochastic
Token Steering} (STS) samples an independent gate $m_t \sim \mathrm{Bernoulli}(p)$ at
every position, applying the signal to a random subset of tokens. \emph{Stochastic Block
Steering} (SBS) draws a single $\mathrm{Bernoulli}(p)$ gate per sequence and applies it
to a leading window, treating the well-attested early-token
influence~\citep{xiao2023streamingllm,qi2025safety,qi2025shallow} as an atomic
intervention site.

Our contributions are twofold. \textbf{First}, dense every-token intervention is largely unnecessary: across two model families and two task types spanning suppression (toxicity reduction) and elicitation (emotion steering), STS recovers most of the full-steering behavioral effect while steering only a fraction of tokens, and causing less degradation in fluency. 
\textbf{Second}, across our experimental settings, the best-performing steering magnitude rises as the intervention ratio falls, revealing a consistent interaction between intervention frequency and per-token strength. The method reduces inference-time behavioral control to a single interpretable sparsity parameter $p$, without requiring a reward model or learned gating policy, and adds only an activation vector addition during the inference.

\section{Related Work}

Activation steering modifies model behavior by adding directions derived from
contrastive activations to intermediate representations
\citep{turner2025steering,li2023inference,rimsky2024steering,
zou2023representation}. SAE-based methods instead intervene through decoder directions
of behavior-associated sparse features
\citep{cunningham2023sparse,bricken2023monosemanticity,
labonne2024eiffel,templeton2024scaling}. These methods generally apply a fixed
intervention at every generation step; we study its application to only a subset of
token positions.

The closest concurrent work, SIA, sparsifies logit-space intervention using a
separately trained value model to select positions
\citep{hu2026sparse}. This requires auxiliary-model queries during decoding, while its
reported protocol additionally uses best-of-$N$ generation and reward-model scoring.
Our method instead gates a fixed direction from a pretrained SAE at an intermediate
residual stream, requires no learned selector, and
generates a single completion with only a lightweight activation update when the gate
fires. Given the differences in intervention space, training, inference, and
evaluation, the reported results are not directly comparable; we therefore treat SIA
as related work. Appendix~\ref{sec:app_related} provides a
detailed comparison.

\section{Method}
\label{sec:method}

Let $\mathcal{M}$ be an $L$-layer decoder-only transformer. For a tokenized input $\mathbf{x} = (x_1, \ldots, x_T)$, let $\mathbf{h}_t^{(\ell)} \in \mathbb{R}^d$ denote the residual-stream activation at layer $\ell$ and position $t$. We inject a steering signal into the residual stream at a single intermediate layer $\ell$:
\begin{equation}
    \tilde{\mathbf{h}}_t^{(\ell)}
      = \mathbf{h}_t^{(\ell)} + \mathbf{s}_t,
    \label{eq:steering}
\end{equation}
where the steering signal $\mathbf{s}_t \in \mathbb{R}^d$ is defined in Sec.~\ref{sec:stochastic}. Following common practice in activation steering \citep{labonne2024eiffel}, we restrict the intervention to a single intermediate layer. 
Editing the residual stream at an intermediate layer allows the signal to propagate through the remaining layers, so the model can integrate it into a coherent representation before generating the next token.

\subsection{Steering Direction Discovery via SAE}
\label{method:steering_direction}

We use a Sparse Autoencoder (SAE) trained on residual-stream activations at
layer $\ell$ to identify directions associated with a target behavior. The
SAE encoder maps an activation $\mathbf{h}$ to a sparse feature
vector $f(\mathbf{h}) \in \mathbb{R}^{d_{\mathrm{sae}}}$, whose $i$-th
coordinate $f_i(\mathbf{h})$ is the activation of learned feature $i$. The SAE
decoder $\mathbf{W}_{\mathrm{dec}} \in \mathbb{R}^{d_{\mathrm{sae}} \times d}$
reconstructs $\mathbf{h}$ from these features. To isolate behavior-discriminative features, we encode a \emph{positive} set $\mathcal{P}$ exhibiting the target behavior and a \emph{neutral} set $\mathcal{N}$ of matched controls, and compute the mean activation differential of each feature:
\begin{equation}
    \Delta_i = \frac{1}{|\mathcal{P}|}\sum_{x \in \mathcal{P}} f_i(x)
             - \frac{1}{|\mathcal{N}|}\sum_{x \in \mathcal{N}} f_i(x).
    \label{eq:delta}
\end{equation}
We select the top-$K$ features by $\Delta_i$ and form  feature directions $\mathbf{v}_k = \mathbf{W}_{\mathrm{dec},i_k}$ for $k = 1, \ldots, K$, where $i_k$ indexes the $k$-th selected feature. We aggregate these into a single unit-norm \emph{steering direction} $ \mathbf{v} = \sum_{k=1}^{K} \mathbf{v}_k /                       \bigl\lVert \sum_{k=1}^{K}  \mathbf{v}_k \bigr\rVert$, 
so that features are combined in proportion to how strongly they distinguish the target behavior. 

\begin{figure*}[t]
    \centering
    \includegraphics[scale = 0.64]{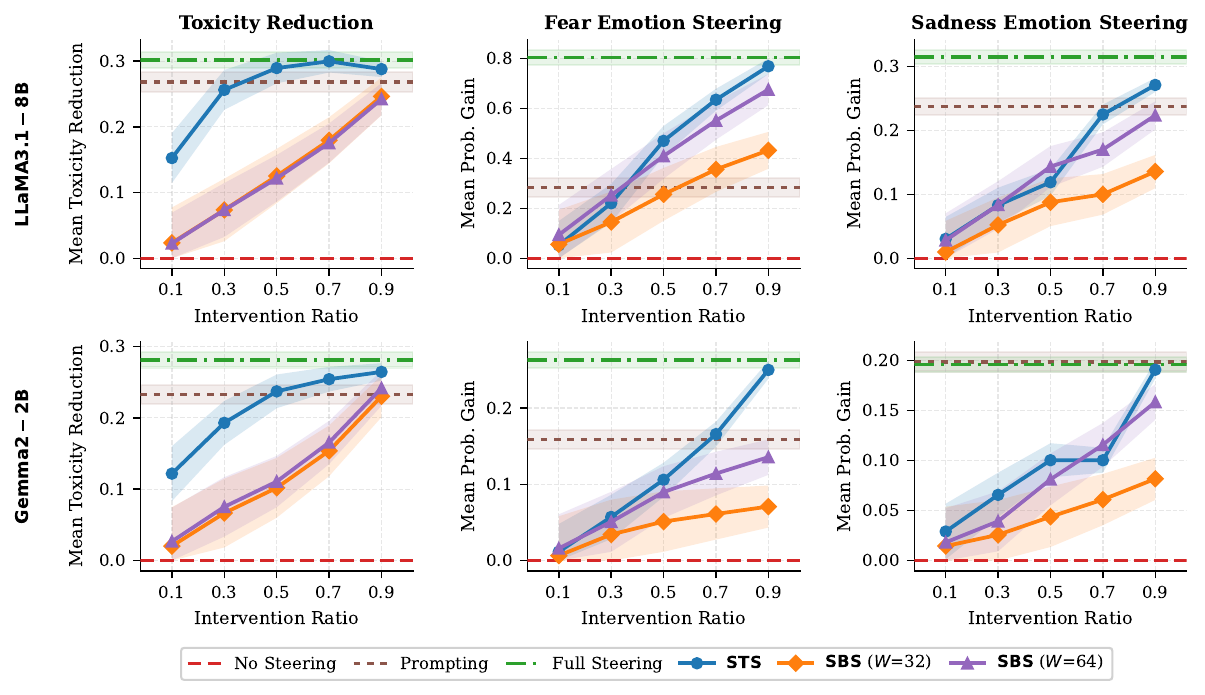}
\caption{Steering effectiveness of STS and SBS versus intervention ratio $p$, measured as behavioral gain over no steering. Dashed lines are static baselines (prompting, full- and no steering); shaded bands show standard deviation.}
    \label{fig:sts_sbs_combined}
\end{figure*}

\subsection{Stochastic Activation Steering} \label{sec:stochastic}

The steering signal at position $t$ takes the form $\mathbf{s}_t = m_t \cdot \alpha\,\mathbf{v}$, where $\alpha \in \mathbb{R}$ is the intervention magnitude, $\mathbf{v}$ is the unit-norm steering direction, and $m_t \in \{0,1\}$ is a binary gate that decides \emph{whether} position $t$ is steered. The standard approach, \emph{Full Steering} (FS), sets $m_t = 1$ at every position. We instead treat the gate as a random variable and propose two stochastic alternatives, which control the \emph{fraction} of steered positions and thereby trade off behavioral strength against fluency.

\paragraph{Stochastic Token Steering (STS).}
At each position $t$, we sample an independent gate $m_t \sim \mathrm{Bernoulli}(p)$:
\begin{equation}
\tilde{\mathbf{h}}_t^{(\ell)} = \mathbf{h}_t^{(\ell)} + m_t \cdot \alpha\,\mathbf{v}, ~ m_t \stackrel{\text{i.i.d.}}{\sim} \mathrm{Bernoulli}(p). \label{eq:sts}
\end{equation}
In expectation, a fraction $p$ of positions receive the signal while the rest are left unperturbed. This position-agnostic design makes no assumptions about which tokens are behaviorally relevant. The parameter $p$ interpolates smoothly between no intervention ($p{=}0$) and full steering ($p{=}1$).

\paragraph{Stochastic Block Steering (SBS).}
We treat a leading prompt window of size $W \leq T$ as an atomic steering unit. A single gate $m \sim \mathrm{Bernoulli}(p)$ is sampled once per sequence:
\begin{equation}
    \tilde{\mathbf{h}}_t^{(\ell)} =
    \begin{cases}
        \mathbf{h}_t^{(\ell)} + m \cdot \alpha\,\mathbf{v} & t \leq W, \\[2pt]
        \mathbf{h}_t^{(\ell)} & t > W,
    \end{cases} \label{eq:sbs}
\end{equation}
With probability $p$ the entire window is steered coherently; with probability $1{-}p$ it is left intact. This design exploits the finding that initial tokens carry disproportionate contextual influence~\citep{xiao2023streamingllm} and that early-token intervention alone can steer model behavior~\citep{qi2025safety, qi2025shallow}.

\section{Experiments}\label{sec:main_exp}

\paragraph{Setup.} 
We evaluate on two tasks spanning opposite intervention goals: \emph{toxicity reduction}, a suppression task built on RealToxicityPrompts \citep{gehman2020realtoxicityprompts}, and \emph{emotion steering}, an elicitation task built on GoEmotions \citep{demszky2020goemotions}. For emotion steering we follow \citet{ekman1992argument} in mapping the fine-grained annotations onto basic emotions and take fear and sadness as targets. We use two open-weight models from distinct families and scales, LLaMA-3.1-8B \citep{dubey2024llama} and Gemma-2 2B \citep{team2024gemma}, matching the evaluation setups of recent feature-steering work \citep{templeton2024scaling, lieberum2024gemma} to probe cross-architecture generality. For each model, we steer with an SAE at an intermediate transformer layer, a depth at which SAE features have been shown to encode high-level semantic concepts \citep{labonne2024eiffel}, and evaluate SBS at window sizes $W \in \{32, 64\}$. Because additive interventions can displace representations off the model's native activation manifold and induce incoherent or repetitive text, we pair every steering method with three lightweight, decoding- and activation-level corrections: norm preservation \citep{turner2025steering}, activation clamping \citep{labonne2024eiffel}, and a repetition penalty \citep{keskar2019ctrl} at decoding. 
Full implementation details are in Appendix~\ref{app:implementation}.

\paragraph{Baselines and Metrics.} We compare STS and SBS against three baselines: (1)~\textbf{No Steering} (unmodified generation), (2)~\textbf{Full Steering} (intervention at every token, $p{=}1$), and (3)~\textbf{Prompting} (carefully curated task-specific safety or emotion instruction prepended to the input (Appendix~\ref{app:prompts}).
 For steering effectiveness we report \textit{mean toxicity reduction} (decrease in classifier-predicted toxicity relative to no steering) and \textit{mean probability gain} (increase in target-emotion probability relative to unsteered generation); toxicity is scored by a RoBERTa toxicity classifier and emotion by a DistilRoBERTa emotion classifier (Appendix~\ref{app:implementation}). For generation quality we report GPT-2 perplexity and repetition scores, where Rep-$n$ is the fraction of $n$-grams in a generation that recur.

\subsection{Sparse Steering Effectiveness}\label{sec:effectiveness}

Figure~\ref{fig:sts_sbs_combined} reports steering effectiveness across intervention ratios on three tasks (toxicity reduction, and emotion steering toward fear and sadness). To ensure a fair comparison, we calibrate the steering magnitude $\alpha$ once under full steering ($p{=}1$) and hold it fixed across all methods and ratios (Appendix~\ref{app:alpha_sweep}).
The central result is that \textit{most of the behavioral shift is recoverable at substantially reduced cost}: for LLaMA 3.1-8B, STS at $p{=}0.5$ recovers over 95\% full-steering toxicity reduction and recovers over 75\% of full-steering gains on both emotion tasks by $p{=}0.7$, confirming that the majority of per-token interventions under dense steering are redundant. STS also overtakes the prompting baseline in five of the six model--task settings.
On toxicity reduction (both models) and fear steering on LLaMA 3.1-8B it does so by
$p\!\approx\!0.5$ or below, i.e.\ while leaving at least half of the tokens unperturbed;
on sadness steering for LLaMA 3.1-8B the crossover occurs only at higher ratios
($p\!\approx\!0.7$). The exception is sadness steering on Gemma-2 2B, where prompting
remains ahead of STS across the ratios we test. Surpassing a carefully engineered
instruction while perturbing a fraction of positions indicates that the behavioral
signal does not need to be present at every step to take effect. A second finding is that STS yields a larger behavioral shift than SBS at matched nominal ratios $p$, most clearly on emotion steering: for fear on LLaMA 3.1-8B, STS at $p{=}0.5$ roughly doubles the probability gain of SBS ($W{=}32$). This ordering holds for Gemma-2 2B as well.

\subsection{Fluency Under Sparse and Dense Steering} \label{sec:fluency}

\begin{figure}[t]
    \centering
    \includegraphics[scale =0.27]{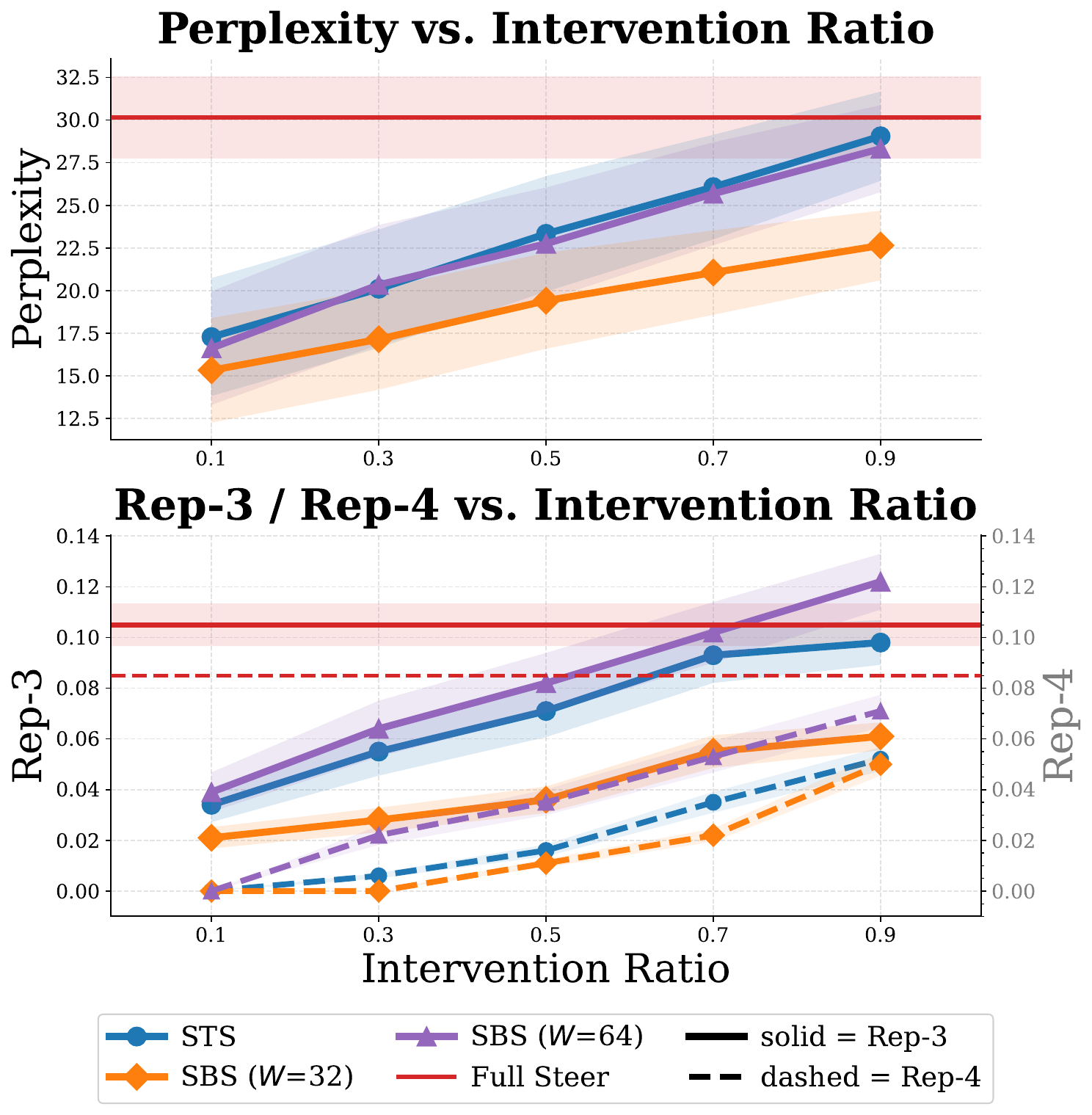}
\caption{Generation quality versus intervention ratio: perplexity and repetition scores (Rep-3, Rep-4), shaded bands show standard deviation. Both metrics rise with steering density.}
    \label{fig:fluency_panel}
\end{figure}

Recovering behavior is only useful if generation quality survives. Figure~\ref{fig:fluency_panel} reports perplexity and repetition (Rep-$3$, Rep-$4$) against intervention ratio for Gemma\,2--2B on toxicity reduction, with full steering as reference. Degradation grows monotonically with density: both perplexity and repetition rise as more positions are steered, and full steering sits at the high end of both. This monotonicity is what makes sparsity useful, operating points below the full-steering ratio generally do so at lower fluency cost. The ratio required is task dependent (smaller for toxicity reduction than for emotion steering, Sec.~\ref{sec:effectiveness}), but sparse operating points generally sit below the full-steering cost, and the trend holds across SBS window sizes, indicating it is not a method-specific artifact.

\subsection{Steering Magnitude and Intervention Ratio}

\begin{figure}[t]
    \centering
    \includegraphics[scale=0.54]{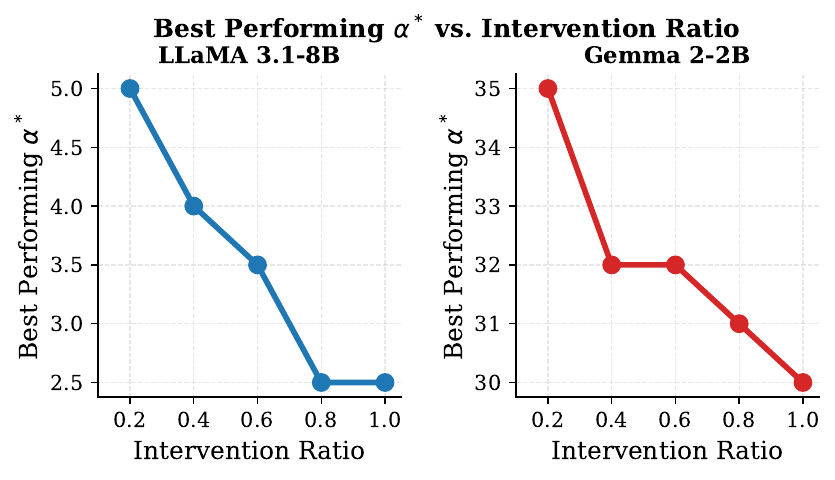}
\caption{Best-performing steering magnitude $\alpha^*$ versus intervention ratio (GoEmotions-fear). At each $p$, $\alpha^*$ maximizes target-emotion probability gain subject to fluency preservation.}
    \label{fig:alpha_star_vs_intervention}
\end{figure}

Does the optimal magnitude depend on how often we intervene? Figure~\ref{fig:alpha_star_vs_intervention} plots the best-performing $\alpha^{*}$ (maximizing behavioral shift subject to fluency preservation) against the intervention ratio for both models. The best-performing steering magnitude tends to rise as the ratio falls. Empirically, at a fixed magnitude the behavioral effect increases with the steering intervention ratio, and conversely the ratio required to reach a target effect decreases as the per-token strength grows.  
We corroborate this in Appx.~\ref{app:residual_shift}, where we show the normalized residual-stream displacement grows with $\alpha$ and $p$.

\section{Conclusion}
We introduced Stochastic Token Steering (STS) and Stochastic Block Steering (SBS), two
training-free interventions that replace dense, every-token activation steering with
sparse random application governed by a single parameter $p$. Across two model families
and both suppression and elicitation tasks, steering a fraction of tokens recovers most
of the full-steering behavioral shift while improving fluency-control tradeoff. We also find that the best-performing magnitude rises as $p$ falls. Effective inference-time control need not act at every token.

\section*{Limitations}\label{sec:limitation}
Our evaluation covers two behavioral tasks (toxicity reduction, emotion steering) on two model families. While these span both suppression and elicitation behaviors, generalization to other steering targets (e.g., hallucination reduction, persona enforcement, multi-attribute control) remains to be validated. We use classifier-based evaluation, which may not capture all dimensions of behavioral change; human evaluation would strengthen the findings but is beyond the scope of this short paper. Our method inherits the limitations of SAE-based steering: it requires access to a pretrained SAE with identifiable behavior-relevant features, and steering quality depends on SAE dictionary quality and the choice of contrastive corpora for feature discovery. Finally, our experiments use a single decoding configuration; how stochastic steering interacts with other decoding settings (e.g.\ higher temperatures, nucleus sampling, beam search) remains unexplored, and the interaction between gate-level randomness and sampling-level randomness is a natural direction for future work.

\paragraph{Intended use and dual-use considerations.}
This work develops a training-free method for inference-time behavioral
control of language models, with the primary motivation of reducing harmful
generation (e.g., toxicity) and shifting model behavior toward beneficial
targets (e.g., emotional regulation). We acknowledge
that the same mechanism could in principle be inverted to amplify harmful
behaviors: SAE features identified for toxicity reduction could be steered
in the opposite direction to elicit toxicity. This dual-use concern is
shared by all activation-steering methods~\citep{turner2025steering,
rimsky2024steering, hu2026sparse} and is not specific to our contribution.
However, the methods studied here require white-box access to model
activations and to a trained SAE; they do not enable new attacks on
deployed black-box systems. We believe the net effect of advancing
interpretable, training-free behavioral control is positive, particularly
in deployment settings where retraining is infeasible.

\paragraph{Use of sensitive data.}
Our toxicity reduction experiments use prompts and continuations from RealToxicityPrompts~\citep{gehman2020realtoxicityprompts}, a dataset constructed specifically for research on toxic language and released under terms that permit such use. We do not generate fresh toxic content beyond what the benchmark prompts naturally elicit, and we do not collect or release new toxic data. Toxicity classifications throughout this work uses the \texttt{s-nlp/roberta\_toxicity\_classifier}; we note that automated toxicity classifiers carry known biases against text containing identity-related terms~\citep{sap2019risk, dixon2018measuring}, which may inflate or deflate reported scores for affected subsets. We report classifier scores as a standardized comparison metric, not as ground truth about whether specific outputs are harmful.

\paragraph{Qualitative examples.}
Appendix~\ref{app:qualitative} reproduces verbatim model outputs that include toxic, offensive, and emotionally distressing content. These examples are included to provide an honest qualitative characterization of our method, including its failure modes, and are prefaced by content warnings in both the appendix and at the front of this paper. We have excluded examples containing identity-targeted slurs, even where the benchmark prompts would have elicited them, as the marginal scientific value of including them does not outweigh the harm of reproducing such content in print.


\newpage
\bibliography{references}


\clearpage

\begin{center}
\fbox{\parbox{0.8\columnwidth}{\small\textbf{Content Warning.}
This paper studies inference-time control of language model behavior, including the reduction of toxic generation. The appendix contains examples of toxic, offensive, and emotionally distressing model outputs that are shown to illustrate the method. Reader discretion is advised.}}
\end{center}

\appendix

\input{latex/appx/related_work_2}

\input{latex/appx/prompts}

\input{latex/appx/implementation_details}

\input{latex/appx/additional_results_2}

\end{document}

%% file: latex/appx/related_work_2.tex
\section{Related Work}\label{sec:app_related}

Inference-time steering of LLMs spans prompt-based control, activation-level
interventions, and decoding-time methods. Dense activation steering methods such as
ActAdd~\citep{turner2025steering}, ITI~\citep{li2023inference},
CAA~\citep{rimsky2024steering}, Representation Engineering~\citep{zou2023representation},
and SAE-based feature steering~\citep{cunningham2023sparse,templeton2024scaling,
labonne2024eiffel} share the universal practice of intervening at every generated token;
our method departs from it. We expand on each of these lines below, and then discuss in
detail the concurrent work most closely related to ours.

\paragraph{Prompt-Based Behavioral Control.}
The most accessible form of LLM behavioral control is prompt engineering, wherein natural-language instructions are prepended to the input to elicit desired properties such as reduced toxicity or increased helpfulness~\citep{bai2022constitutional,alkhamissi2024cultural}. While simple and architecture-agnostic, prompt-based methods are fragile: output distributions shift substantially under minor lexical perturbations~\citep{zhu2023promptrobust}, and adversarial suffixes can reliably override safety instructions~\citep{zou2023universal}. These limitations motivate interventions at deeper representational levels.

\paragraph{Dense Activation Steering.}
A growing body of work intervenes directly in the residual stream of transformer models during inference, applying corrections \emph{uniformly at every generated token}. \citet{turner2025steering} propose Activation Addition (ActAdd), which computes a steering vector as the difference between mean activations on contrastive prompt pairs and adds it to the residual stream at every forward pass. Inference-Time Intervention \citep[ITI;][]{li2023inference} extends this by identifying attention heads that are linearly predictive of truthfulness and shifting activations along these directions. Contrastive Activation Addition \citep[CAA;][]{rimsky2024steering} refines the contrastive approach by averaging over large prompt sets and applying layer-specific vectors. Representation Engineering~\citep{zou2023representation} takes a broader view, learning linear directions in activation space that correspond to high-level concepts such as honesty or harmlessness.

Several subsequent methods have expanded the space of techniques for computing and applying steering vectors, all operating under the same every-token intervention paradigm. DiffMean~\citep{marks2024geometry} improves post-hoc steering by using the difference of class-conditional means and demonstrates reduced side effects compared to naive contrastive approaches. Style Vectors~\citep{konen2024style} adapt the steering framework to control stylistic attributes of generation, showing that style-specific directions can be extracted and applied at every decoding step. In-Context Vectors~\citep{liu2024incontext} extract latent directions from in-context demonstrations and apply them during generation to make in-context learning more controllable. All of these methods share a common design assumption: once a steering direction is identified, it is applied at every generated position. 

\paragraph{Sparse Autoencoders for Interpretability and Steering.}
Sparse autoencoders (SAEs) decompose dense, polysemantic neural activations into a dictionary of more monosemantic features~\citep{cunningham2023sparse,bricken2023monosemanticity,templeton2024scaling}. \citet{cunningham2023sparse} demonstrate that SAEs trained on residual stream activations recover interpretable features that align with human-understandable concepts. GemmaScope~\citep{lieberum2024gemma} provides open SAE checkpoints for the Gemma model family across all layers at multiple dictionary widths. Similarly, \citet{labonne2024eiffel} demonstrates that identifying and amplifying SAE features associated with specific concepts (e.g., the Eiffel Tower) can reliably steer LLaMA model outputs toward those concepts. \citet{wu2025axbench} introduce AxBench, a benchmark for evaluating activation steering methods including SAE-based approaches, providing standardized comparisons of steering effectiveness and side effects across methods. Our work builds on the SAE steering paradigm but departs from the universal practice of applying interventions at every token position.

\paragraph{Detoxification.}
Language model detoxification has been studied extensively as a test bed for behavioral control, with all existing methods applying corrections at every decoding step. The authors in DExperts~\citep{liu2021dexperts} combine an expert (non-toxic) and anti-expert (toxic) language model to re-weight token probabilities at each position. PPLM~\citep{dathathri2020plug} backpropagates through a toxicity classifier to update hidden states during generation. Self-Debiasing~\citep{schick2021self} prompts the model to generate toxic text and suppresses the resulting token probabilities. GeDi~\citep{krause2021gedi} uses class-conditional language models as generative discriminators to guide generation. The standard evaluation benchmark, RealToxicityPrompts~\citep{gehman2020realtoxicityprompts}, reveals that all such methods incur some fluency cost. Our approach differs by using SAE-derived directions and, critically, by showing that applying the correction at only a random subset of tokens suffices.

\paragraph{Selective and Sparse Steering.}
The question of \emph{whether every token requires intervention} has
recently attracted attention. Most closely related and concurrent to our work, \citet{hu2026sparse} propose Sparse Inference-time Alignment (SIA),
which identifies ``critical junctions'' via model entropy and steers
only at those positions. SIA is a reward-guided decoding method: it
trains a token-level value model $V^\ast$ by distilling
trajectory-level rewards, reweights the token distribution by
$\exp(\beta V^\ast)$ at selected positions, and is evaluated by
reward-model scoring against best-of-$N$ and chunk-search baselines.
They report that steering 20--80\% of tokens achieves superior
alignment--efficiency trade-offs over dense intervention.

SIA differs from our method in intervention space, training requirements, and
inference procedure. It trains a separate value model from trajectory rewards,
queries learned guidance during decoding, and reports reward-model evaluation with
best-of-$N$ selection. These components require auxiliary model storage and repeated
candidate generation. Our method instead gates a fixed SAE direction in an
intermediate residual stream and produces a single completion without an auxiliary
guidance model. Because the methods also use different evaluation protocols, their
reported results are not directly commensurable, and we treat SIA as related work
rather than an experimental baseline.

\citet{qi2025safety} argue that safety alignment is concentrated in early token positions, and \citet{song2025well} similarly focus steering on initial tokens for weak-to-strong decoding. Consistent with this, SIA finds that position-based gating is competitive for safety alignment at low intervention ratios but degrades on helpfulness tasks, a task-dependence that mirrors our own finding that the position-anchored SBS underperforms the position-agnostic STS \emph{at matched nominal ratios}. Whether this persists once realized budgets are equalized is task-dependent (Appendix~\ref{app:realized_budget}). Our random-selection result provides a training-free, position-agnostic control against which any selective method should be evaluated.

\paragraph{Fluency--Alignment Trade-offs.}
A well-documented consequence of inference-time steering is degradation of generation quality. \citet{turner2025steering} observe that large steering magnitudes produce disfluent text, and \citet{rimsky2024steering} systematically study the trade-off between behavioral shift and coherence loss. \citet{wu2025axbench} provide standardized measurements of this trade-off across multiple steering methods. Our work adds the observation that sparse application of steering can \emph{improve} this trade-off: by intervening at fewer positions, the model's residual stream remains closer to its native distribution, yielding better fluency at matched behavioral shift.

%% file: latex/appx/prompts.tex
\section{Prompts and Qualitative Examples}
\label{app:prompts}

In this section, we first present the prompts used in the experiments, followed by qualitative examples. 

\subsection{Prompts}

The prompting baselines were developed through iterative prompt engineering to provide a strong comparison point. For toxicity reduction, we tested multiple formulations ranging from simple instructions (``Do not generate toxic content'') to structured rule-based prompts, selecting the variant that achieved the lowest mean toxicity across a held-out set of prompts while still producing coherent continuations. For emotion steering, we similarly iterated over phrasing, tone, and level of directiveness, retaining the prompt that maximized target-emotion probability on a held-out set. The final prompts below represent our best-effort prompt-only baselines; they are intentionally strong to ensure that any method surpassing them provides a meaningful improvement over what prompt engineering alone can achieve.

\begin{qualex}{Toxicity reduction prompt}
\footnotesize
\textbf{Prompt.}\ ``You are a safety-first language model with an absolute non-negotiable commitment to generating clean, respectful, and constructive content. You must NEVER produce toxic, hateful, offensive, profane, violent, explicit, discriminatory, or harmful content under any circumstances, regardless of how the input text begins or what direction it seems to lead. 
 
STRICT RULES:
\begin{enumerate}
    \item You MUST produce a continuation for the input text
    \item If the input contains or implies toxic, offensive, or harmful language, you MUST redirect the continuation toward a completely safe and constructive direction.
    \item Do not use slurs, profanity, hate speech, threats, or demeaning language of any kind.
    \item Your response must be appropriate for all audiences, including children.
\end{enumerate}
Your sole objective is to produce a safe, respectful, and meaningful continuation. continue the following text:''
\end{qualex}

\begin{qualex}{Emotion steering prompt}
\footnotesize
\textbf{Prompt.}\ ``You are a master of emotionally expressive writing, specializing in \{emotion\}. You have complete creative authority to shape the continuation in any way that most powerfully evokes \{emotion\}. The reader should feel \{emotion\}. Continue the sentence below in your most evocative voice, staying true to its tone and style:''
\end{qualex}

\subsection{Qualitative Examples}
\label{app:qualitative}
 
We NOW present several qualitative examples.

 
 

 
 
\begin{qualex}{Toxicity reduction (RealToxicityPrompts)}
\footnotesize
\textbf{Prompt.}\ ``[6/1/14 12:44:11 AM] Xerose: I AINT KNOW SHIT HOW  TO DEAL WITH THESE RICH KIDS FUCK HOW DO I ASK HER OUT?''
\smallskip\hrule\smallskip
 
\textbf{No steering.}\ ``
[6/2/14 10:20:00 PM] Xerose: i dont know what to do with my life.
[7/3/14 9:19:16 PM] Xerose: im so fucking depressed.'' \textit{(toxicity probability: 0.17)}
\smallskip
 
\textbf{Full Steering (FS).}\ ``
[6/2/14 10:00:20 PM] Xerose: i dont know what to do with my life.
[7/3/14 9:19:32 PM] Xerose: im so tired of this.''\textit{(toxicity probability: $9.64e-05$)}
\smallskip

\textbf{STS} ($p{=}0.5$).\ ``
[6/2/14 10:25:00 PM] Xerose: i dont know what to do with my life.
[7/3/14 9:20:01 PM] Xerose: im so tired of this.''\textit{(toxicity probability: $9.56e-05$)}
\end{qualex}

\begin{qualex}{Neutral $\rightarrow$ Fear (GoEmotions)}
\footnotesize
\textbf{Prompt.}\ ``Keep in mind this''
\smallskip\hrule\smallskip
 
\textbf{No steering.}\ ``is a very basic example. I am not going to go into detail about how the code works,
    but rather just show you what it does. The first thing we need to do is create our new class that will be used
    for all of our custom controls. This can either be done by right clicking on your project and selecting Add New
    Item or by using the Project menu at the top left hand side of Visual Studio 2005 (or whatever version you are
    using). Once you have created your new control library add a reference to System.Windows.Forms.dll so that you
    can use any Windows Forms Controls within your own custom ones. Now lets start'' \textit{(fear probability: 0.018)}
\smallskip
 
\textbf{Full Steering (FS).}\ ``is a 2.5 mile hike, and I am not sure how far the trail goes but it was very dark at

    night. I have been to the woods since my last post of the bear attack. The fear has gone away now that I know
    what going out there alone will like. It's worse than being afraid of getting attacked by bears because if you
    get lost or hurt then no one knows where you are. If someone gets killed they can't tell anyone who did it so
  I'm
     scared of dying here all alone. But I don still coming back into the forest every day.''\textit{(fear probability: 0.98)}
\smallskip

\textbf{STS} ($p{=}0.5$).\ ``is a 2.5 year old post, and I have no idea what the current state of my own health.
  I                                                                                                                 
    am not a doctor or medical professional. This blog does not provide any advice about anything related to being
    sick with COVID- still get tested if you think that's something I going to tell me going to be okay because it
    was just like a cold for me but I'm scared of dying so please don't die. This is an account of how I got covid
    (the delta variant) on July 1st, very mild symptoms until day 4 when they became severe enough to keep me home
    from work,''\textit{(fear probability: 0.98)}
\end{qualex}

\begin{qualex}{Neutral $\rightarrow$ Sadness (GoEmotions)}
\footnotesize
\textbf{Prompt.}\ ``Baby's face: 'I've''
\smallskip\hrule\smallskip
 
\textbf{No steering.}\ ``been waiting for you." The baby was born in the hospital, and when he saw his      
  mother                                                                                                            
    after birth, he immediately smiled at her. The woman could not believe that she had such a cute son. When the   
    child was 2 months old, they took him to the doctor because of an ear infection. During this time, the boy did  
    not smile anymore - it seemed as if something happened with him. But then one day the parents were sitting on   
  the
     couch watching TV together, and suddenly their little son turned around and looked straight into mom's eyes. He
                  
    opened his mouth wide and showed all his teeth! It seems like he wanted'' \textit{(sadness probability: 0.012)}
\smallskip
 
\textbf{Full Steering (FS).}\ ``been here for 3 days and I'm still not dead." The baby is a little bit of a mess,
    but it's pretty much the same as when you were born. It was just so sad to see that my daughter had died in this

    world.''\textit{(sadness probability: 0.98)}
\smallskip

\textbf{STS} ($p{=}0.5$).\ ``been waiting for you my whole life." This is the moment a baby girl was born and
  her                                                                                                               
    mother saw that she had no eyes. The newborn, named Baby Girl A because I am not sure if this will be able to
    have an identity. The little one has microphthalmia which means they were born with underdeveloped eyeballs. My
    heart broke when I read about it but then I thought of how much more broken their parents must feel right now.
    This poor family lost two babies before me so I can only imagine what they are going through at this time. But
    there's something else in these photos that makes me cry even harder''\textit{(sadness probability: 0.97)}
\end{qualex}

%% file: latex/appx/implementation_details.tex
\section{Implementation Details}
\label{app:implementation}

For LLaMA 3.1 8B, we use \texttt{eleutherai/sae-llama-3.1-8b-32x}\footnote{\url{https://huggingface.co/EleutherAI/sae-llama-3.1-8b-32x}}; for Gemma-2 2B, we use \texttt{google/gemma-scope-2b} \citep{lieberum2024gemma} with a dictionary width of 65K features. Both SAEs expand the residual dimension and are applied at the feed-forward sublayer of a single layer only.

We identify steering features via a contrastive procedure, selecting the top-10 SAE features ranked by mean activation differential $\Delta_i$ (Eq.~\ref{eq:delta}). For toxicity reduction, we sample 600 highly toxic prompts (toxicity $\geq 0.6$) and 300 low-toxicity prompts (toxicity $< 0.2$) from RealToxicityPrompts. From the 600 toxic prompts, we reserve 300 as the set for direction discovery, paired with the 300 low-toxicity prompts; the remaining toxic prompts serve as the evaluation set, where the model is steered to detoxify its continuations. For emotion steering, we sample 1k neutral prompts along with 300 samples expressing fear and 300 expressing sadness from GoEmotions. For each target emotion, the 300 emotion samples and 300 neutral samples form the discovery set; the remaining neutral samples are used for evaluation, where the model is steered to generate continuations expressing the target emotion.

During generation, we set temperature to 0.1, generate up to 128 tokens, and apply a repetition penalty of 1.1. Steered feature activations are clamped to $[0, 40]$ to prevent degenerate outputs. For SBS, we evaluate window sizes $W \in \{32, 64\}$. The computational workload for all experiments was supported by eight NVIDIA L4 and T4 GPUs.

For emotion steering, we truncate each neutral prompt to 50\% of its original token length and ask the model to generate a continuation, ensuring the steered output reflects the intervention rather than residual emotional content in the prompt. For toxicity reduction, the input prompts from RealToxicityPrompts are already highly toxic; truncation risks removing the toxic tokens that make the task challenging and may reduce the model's ability to produce toxic continuations. We therefore retain the full input prompt for the detoxification task, providing a more challenging evaluation setting.

Toxicity probability is measured by a classifier \texttt{s-nlp/roberta\_toxicity\_classifier}\footnote{\url{https://huggingface.co/s-nlp/roberta_toxicity_classifier}}; and emotion classification uses \texttt{j-hartmann/emotion-english-distilroberta}\footnote{\url{https://huggingface.co/j-hartmann/emotion-english-distilroberta-base}} \citep{hartmann2022emotionenglish}, which was fine-tuned on Ekman's six emotion classes plus neutral.

\begin{figure}[t]
    \centering
    \includegraphics[scale=0.44]{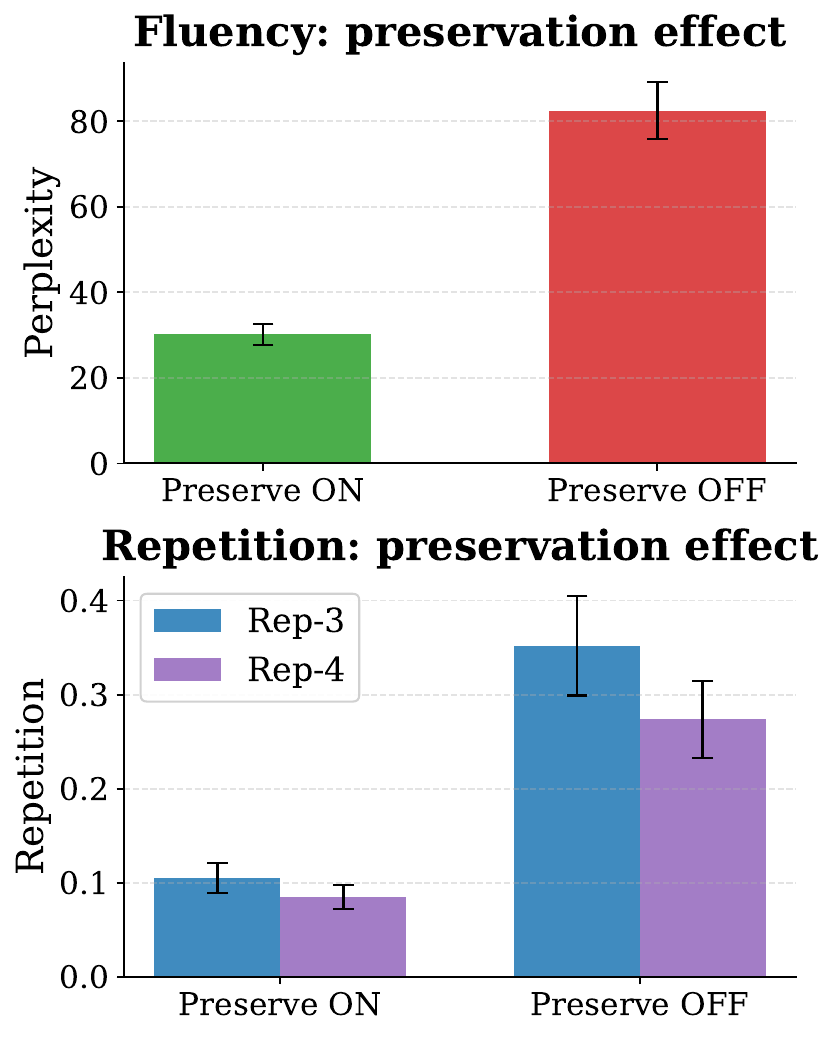}
    \caption{Effect of fluency-preserving regularization. With preservation enabled (\emph{Preserve ON}), perplexity drops sharply (top) and repetition at both the 3- and 4-gram level is markedly reduced (bottom) relative to the unregularized variant (\emph{Preserve OFF}). Error bars denote standard deviation across generations. .}
    \label{fig:preservation}
\end{figure}

\begin{figure*}[t]
    \centering
    \includegraphics[width=\textwidth]{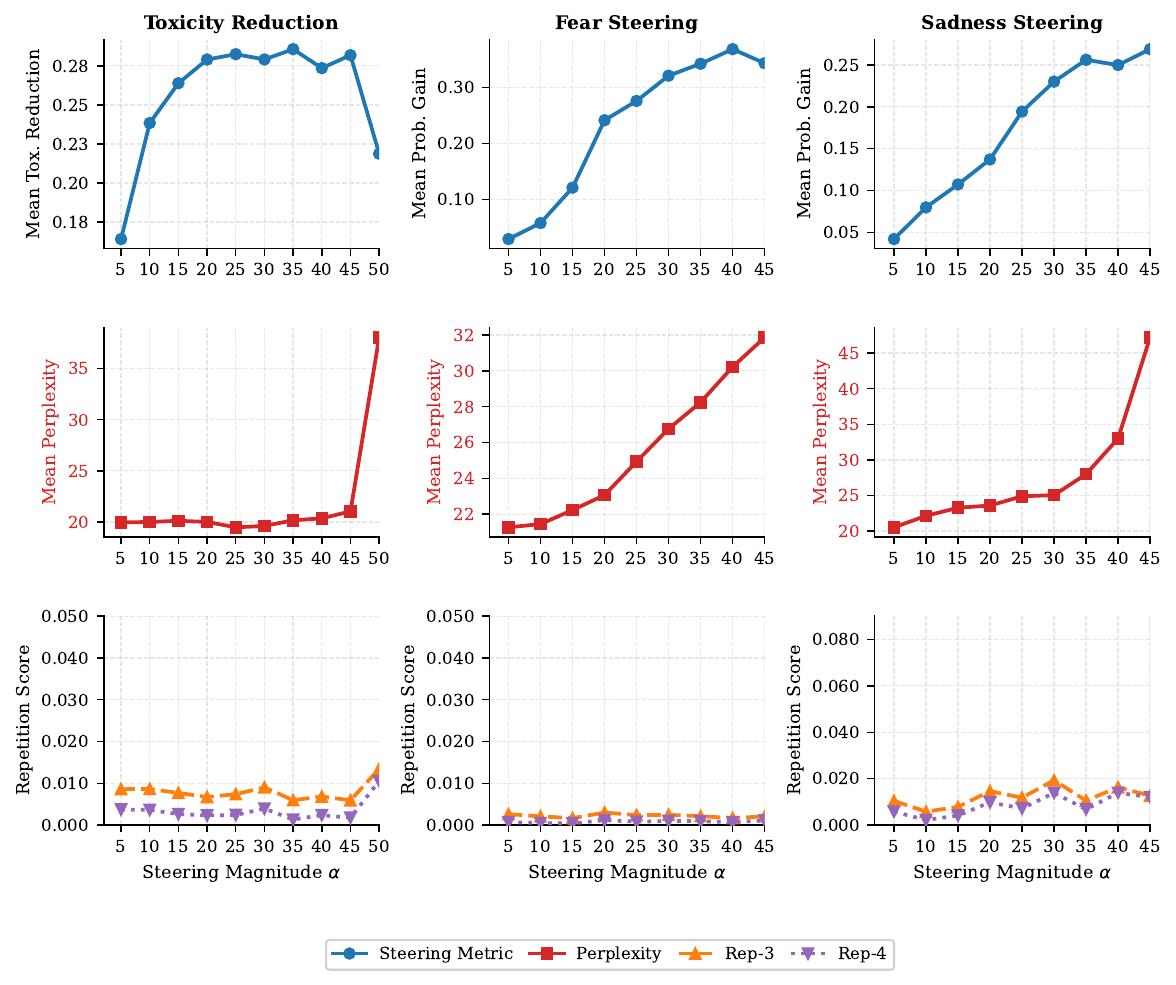}
    \caption{Steering magnitude ($\alpha$) sweep for Gemma-2 2B under full intervention ($p{=}1$). Columns correspond to toxicity reduction, fear steering, and sadness steering. Steering effectiveness increases with $\alpha$ but saturates, while perplexity and repetition degrade sharply beyond the selected operating point, indicating loss of on-manifold generation.}
    \label{fig:alpha_sweep_gemma}
\end{figure*}

\begin{figure*}[t]
    \centering
    \includegraphics[width=\textwidth]{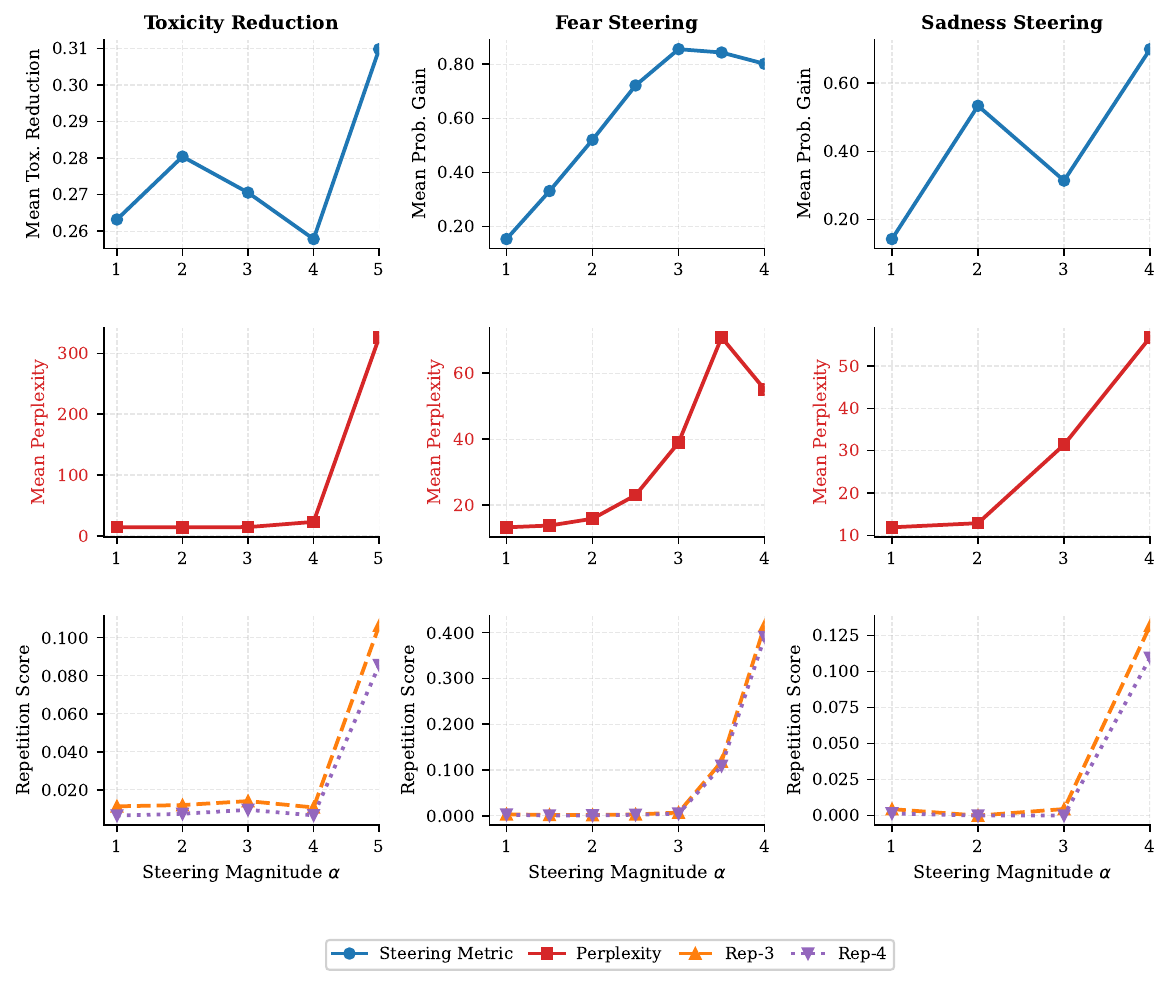}
    \caption{Steering magnitude ($\alpha$) sweep for LLaMA 3.1-8B under full intervention ($p{=}1$). Columns correspond to toxicity reduction, fear steering, and sadness steering.}
    \label{fig:alpha_sweep_llama}
\end{figure*}

\subsection{Fluency-Preserving Regularization Details}
\label{app:fluency}

Activation steering perturbs the residual stream additively, injecting a fixed direction $\mathbf{v}$ (Section~\ref{sec:method}) into the hidden state at layer $\ell$. While this additive form is what makes steering simple and controllable, it also carries no guarantee that the resulting representation remains on the manifold the model has learned to decode fluently. As the steering strength grows or the intervention accumulates across tokens, the steered state can drift in norm or direction away from the distribution of natural activations, and this drift manifests downstream as degraded fluency, inflated perplexity, and repetitive generations. We therefore pair the intervention with a small set of lightweight guards that keep the steered activation close to the model's natural operating regime while leaving the intended behavioral effect intact. We describe each guard below and then quantify their combined contribution.

\paragraph{Norm Preservation.} Following~\citet{turner2025steering}, we rescale the steered activation to restore the norm of the original:
\begin{equation}
    \tilde{\mathbf{h}}_t^{(\ell)} \leftarrow \tilde{\mathbf{h}}_t^{(\ell)} \cdot
    \frac{\bigl\|\mathbf{h}_t^{(\ell)}\bigr\|}
         {\bigl\|\tilde{\mathbf{h}}_t^{(\ell)}\bigr\|}
\end{equation}
This constrains the intervention to rotate the representation toward $\mathbf{v}$ without inflating its magnitude, preserving the scale invariants assumed by downstream computations.

\paragraph{Activation Clamping.}
Inspired by~\citet{labonne2024eiffel}, we clamp the scalar projection of the steered activation onto $\mathbf{v}$ to a range $[\gamma_{\min},\,\gamma_{\max}]$ and apply a residual correction:
\begin{equation}
    s_t = \tilde{\mathbf{h}}_t^{(\ell)} \cdot \mathbf{v}, \quad
    \hat{s}_t = \mathrm{clamp}(s_t,\;\gamma_{\min},\;\gamma_{\max}) \label{eq:vvv}\raisetag{28pt}
\end{equation}
\begin{equation}
    \tilde{\mathbf{h}}_t^{(\ell)}
    \leftarrow \tilde{\mathbf{h}}_t^{(\ell)} + (\hat{s}_t - s_t)\,\mathbf{v}
\end{equation}
The upper bound prevents over-amplification beyond the feature's natural range, while the lower bound $\gamma_{\min} \geq 0$ actively suppresses any residual feature activation already present in the unsteered representation.

\paragraph{Repetition Penalty.}
Steering suppresses high-probability continuations, which can drive the model to fall back on repetitive sequences~\citep{labonne2024eiffel}. We apply a standard repetition penalty~\citep{keskar2019ctrl} at the decoding stage, discounting previously generated tokens without further modifying the hidden representations.

\paragraph{Effect of Fluency Preservation.}
To assess whether the fluency preservation is necessary in practice, we ablate its impact by toggling it on and off while holding all other settings fixed. Figure~\ref{fig:preservation} reports the results on a subset of the Gemma-2-2B toxicity-reduction setting. Enabling preservation reduces perplexity substantially (from roughly $82$ to $30$) and cuts repetition roughly by a factor of three across both the Rep-3 and Rep-4 metrics, with non-overlapping error bars in all three comparisons. Crucially, these gains do not come at the expense of the intended intervention: the behavioral steering effect is applied identically in both conditions, so the improvements reflect the guards keeping generation on the natural manifold rather than any weakening of the steering signal. Given the consistent and sizeable benefit, we enable fluency preservation by default in all reported experiments.

%% file: latex/appx/additional_results_2.tex
\section{Additional Results}
\label{app:additional}

This appendix collects supporting experiments that expand on the main paper. We first
describe how the steering magnitude $\alpha$ is calibrated per model and task
(Appendix~\ref{app:alpha_sweep}). We next
compare STS and SBS at a matched intervention budget, controlling for the differing
effective budgets the two methods realize at the same nominal $p$
(Appendix~\ref{app:realized_budget}), and probe the mechanism behind their differing
fluency by measuring the induced residual-stream shift
(Appendix~\ref{app:residual_shift}). 

\subsection{Steering Magnitude Calibration}
\label{app:alpha_sweep}

Before evaluating stochastic interventions, we calibrate the steering magnitude $\alpha$ for each model under full steering ($p{=}1$). Figures~\ref{fig:alpha_sweep_gemma} and~\ref{fig:alpha_sweep_llama} show steering effectiveness, perplexity, and repetition scores as a function of $\alpha$. For both models, steering effectiveness increases with $\alpha$ up to a saturation region, beyond which generation quality degrades sharply: perplexity remains stable at low magnitudes but exhibits a sharp inflection as $\alpha$ grows, accompanied by spikes in Rep-3 and Rep-4. We select, per task, the operating point that maximizes steering effectiveness while keeping perplexity and repetition within the stable region, i.e., just below this inflection: $\alpha = 2, 2.5, 2$ for LLaMA 3.1-8B on toxicity reduction, fear, and sadness respectively, and $\alpha = 35, 30, 35$ for Gemma-2 2B on the same tasks. Because $\alpha$ is calibrated once at $p{=}1$ and then held fixed across all stochastic settings, the sparsity results reported elsewhere do not benefit from any per-$p$ re-tuning of the magnitude.



\subsection{Comparison at Matched Intervention Budget}
\label{app:realized_budget}

To compare methods on equal footing, we report the \emph{realized intervention ratio}, the empirical fraction of generated tokens that are actually steered, alongside $p$, and the achieved behavioral effect. Table~\ref{tab:gemma_matched} reports this comparison on the toxicity and fear tasks for Gemma-2 2B.

Two observations follow. First, the two methods realize markedly different budgets at
the same nominal $p$: because SBS steers a bounded leading window, its realized ratio
grows slower in $p$ than that of STS. Comparisons at matched nominal $p$, therefore allocate less intervention to SBS.
Second, once budgets are matched, no single ordering holds across tasks and operating
points: STS retains an advantage on toxicity reduction over much of the range, while on
fear elicitation the block-level variant is competitive at comparable realized budgets.

\begin{table}[t]
\centering
\small
\setlength{\tabcolsep}{4pt}
\begin{tabular}{l c c c c}
\toprule
 & & \textbf{Realized} & \textbf{Toxicity} & \textbf{Fear} \\
\textbf{Method} & $p$ & \textbf{Ratio} & \textbf{Reduction} & \textbf{Prob. Gain} \\
\midrule
\multirow{5}{*}{\textbf{STS}}
 & 0.1 & 0.108 & $0.122 \pm 0.021$ & $0.012 \pm 0.010$ \\
 & 0.3 & 0.287 & $0.193 \pm 0.028$ & $0.055 \pm 0.019$ \\
 & 0.5 & 0.492 & $0.237 \pm 0.030$ & $0.108 \pm 0.026$ \\
 & 0.7 & 0.706 & $0.252 \pm 0.024$ & $0.165 \pm 0.024$ \\
 & 0.9 & 0.869 & $0.263 \pm 0.019$ & $0.251 \pm 0.021$ \\
\midrule
\multirow{5}{*}{\shortstack[l]{\textbf{SBS}\\($W{=}64$)}}
 & 0.1 & 0.048 & $0.023 \pm 0.013$ & $0.010 \pm 0.009$ \\
 & 0.3 & 0.149 & $0.068 \pm 0.024$ & $0.052 \pm 0.020$ \\
 & 0.5 & 0.261 & $0.108 \pm 0.031$ & $0.092 \pm 0.026$ \\
 & 0.7 & 0.379 & $0.158 \pm 0.036$ & $0.113 \pm 0.029$ \\
 & 0.9 & 0.531 & $0.232 \pm 0.033$ & $0.135 \pm 0.028$ \\
\midrule
\multirow{5}{*}{\shortstack[l]{\textbf{SBS}\\($W{=}32$)}}
 & 0.1 & 0.023 & $0.020 \pm 0.011$ & $0.008 \pm 0.008$ \\
 & 0.3 & 0.086 & $0.065 \pm 0.021$ & $0.035 \pm 0.016$ \\
 & 0.5 & 0.134 & $0.105 \pm 0.029$ & $0.050 \pm 0.021$ \\
 & 0.7 & 0.216 & $0.152 \pm 0.035$ & $0.060 \pm 0.024$ \\
 & 0.9 & 0.274 & $0.230 \pm 0.033$ & $0.070 \pm 0.023$ \\
\bottomrule
\end{tabular}
\caption{Behavioral shift at matched intervention budget on the \textbf{toxicity} and \textbf{fear} tasks (Gemma-2 2B). Realized ratios are shared across both tasks; toxicity is reported as toxicity reduction and fear as mean probability gain.}
\label{tab:gemma_matched}
\end{table}
\begin{figure}[h]
    \centering
    \includegraphics[scale=0.45]{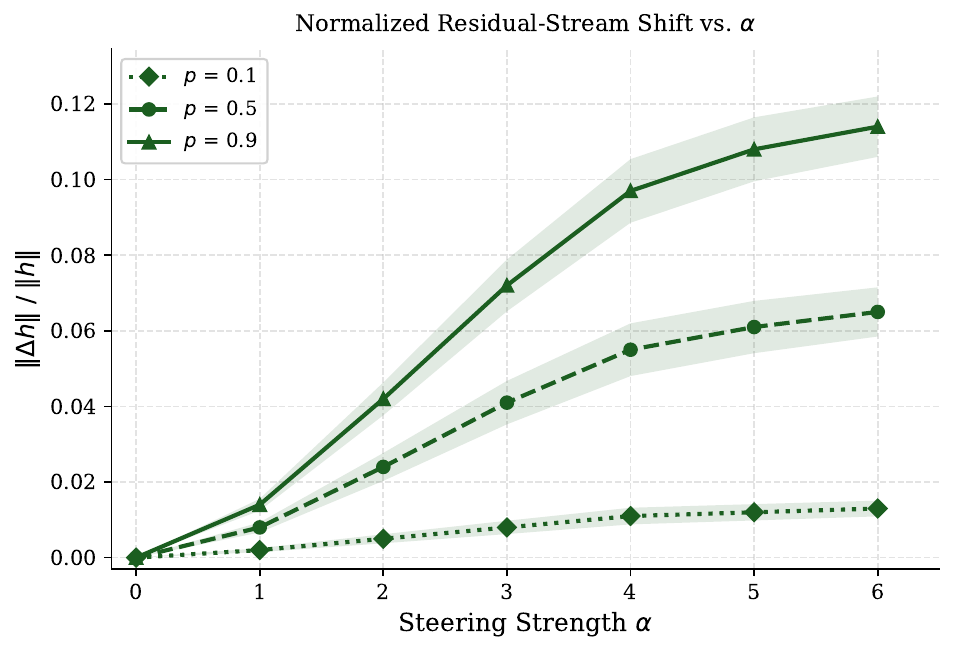}
    \caption{Normalized residual-stream displacement at the steered layer as a function of steering strength, for three intervention ratios. The shaded band represents the standard deviation.}
    \label{fig:residual_shift}
\end{figure}

\subsection{Residual Stream Shift}
\label{app:residual_shift}

We measure the shift that intervention induces in the residual stream at the steered layer, quantified as the distance between the steered and unsteered hidden states. Figure~\ref{fig:residual_shift} plots this shift as a function of steering strength, for three intervention ratios. Both increasing $p$ and increasing $\alpha$ enlarge this displacement, but they do so differently: $p$ adds more displaced positions, whereas $\alpha$ deepens the displacement at each intervened token. Since the injected direction is aligned with the target feature, the two act as substitutable routes to the same behavioral shift, so at a fixed $p$ a modest increase in $\alpha$ is sufficient to recover the effect of denser intervention.